\documentclass[sigconf,nonacm]{acmart}
\usepackage{times}
\usepackage{epsfig}
\usepackage{graphicx}
\usepackage{amsmath}
\usepackage{amssymb}
\usepackage{multirow}

\AtBeginDocument{%
  \providecommand\BibTeX{{%
    \normalfont B\kern-0.5em{\scshape i\kern-0.25em b}\kern-0.8em\TeX}}}

\acmSubmissionID{71}

\begin{document}

\title{No-Reference Image Quality Assessment via Feature Fusion and Multi-Task Learning}

\author{S. Alireza Golestaneh}
\affiliation{%
  \institution{Robotics Institute}
  \institution{Carnegie Mellon University}
  \streetaddress{}
  \city{Pittsburgh}
  \country{USA}}

\author{Kris Kitani}
\affiliation{%
  \institution{Robotics Institute}
  \institution{Carnegie Mellon University}
  \streetaddress{}
  \city{Pittsburgh}
  \country{USA}}



\begin{abstract}
Blind or no-reference image quality assessment (NR-IQA) is a fundamental, unsolved,  and yet challenging problem due to the unavailability of a reference image.
It is vital to the streaming and social media industries that impact billions of viewers daily.
Although previous NR-IQA   methods leveraged different feature extraction approaches,  the performance bottleneck still exists.
In this paper, we propose a simple and yet effective  general-purpose no-reference (NR) image quality assessment (IQA)
framework based on multi-task learning.
Our model employs distortion types as well as subjective human scores to predict image quality.
We propose a feature fusion method to utilize
distortion information to improve the quality score estimation task.
In our experiments, we demonstrate that by utilizing multi-task learning and our proposed feature fusion method,  our model yields better performance for the NR-IQA task.
To demonstrate the effectiveness of our approach, we test our approach on seven standard datasets and show that we achieve state-of-the-art results on various datasets.
\end{abstract}



\settopmatter{printfolios=true}
\maketitle

\section{Introduction}

Image and video compression and applications of visual
media continue to be in high demand these days.
There has been an increasing demand for accurate image and
video quality assessment algorithms for different multimedia and computer vision applications,
such as image/video compression, communication, printing,
display, restoration, segmentation, and fusion \cite{dodge2016understanding,liu2017quality,guo2017building,zhang2017learning}. 
Robustness of different  multimedia and computer vision  applications heavily relies on their input's quality.
Therefore, it is of  great importance to be able to automatically evaluate image quality in the same way as it perceived by the human visual system (HVS).
Furthermore, in many real-world applications,     IQA task needs to be
carried out in a timely fashion on a computationally limited platform.


\begin{figure}[t]
\centering
  \includegraphics [scale=.18]{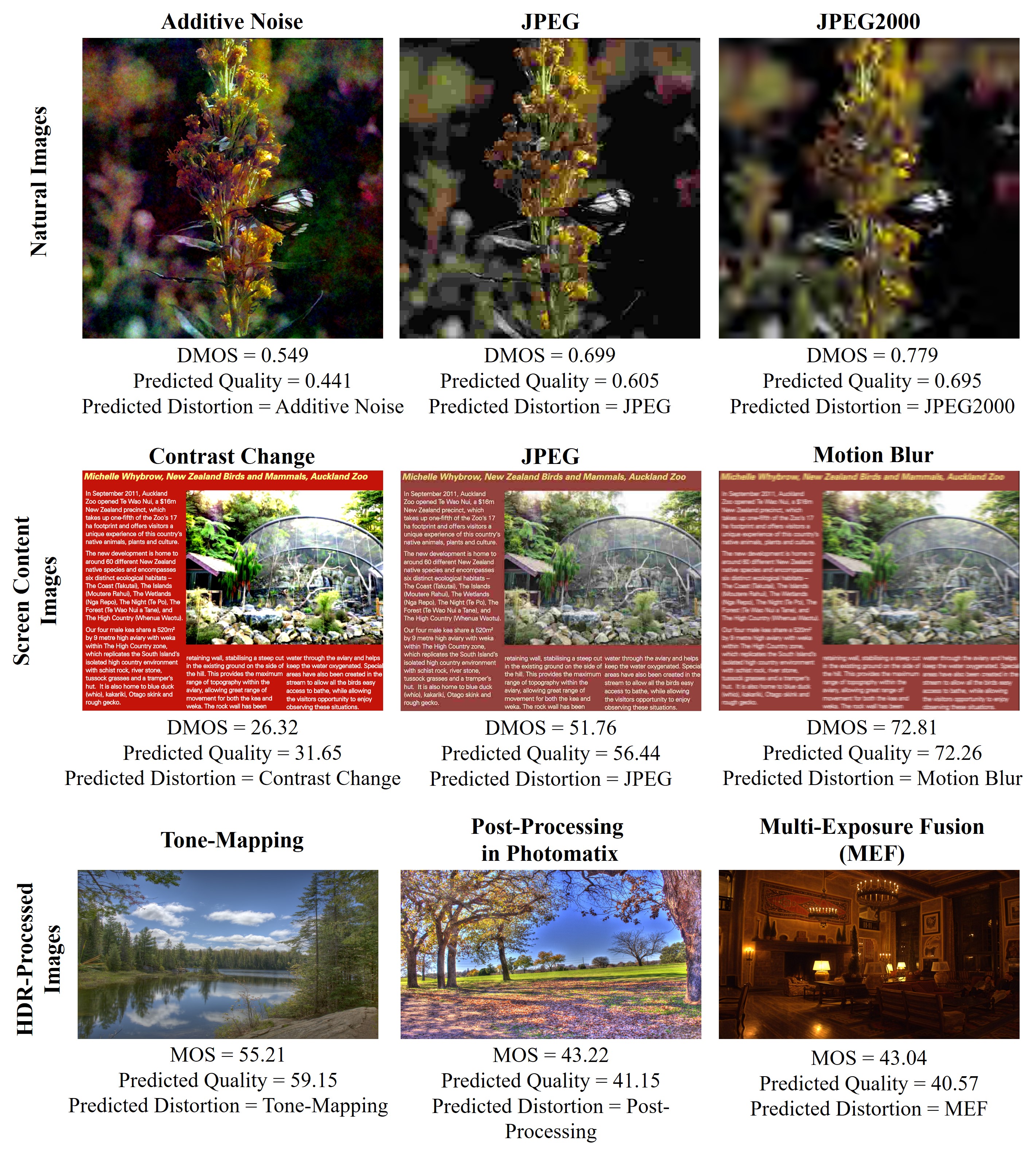}
	\caption{ Example results of our proposed NR-IQA method. 
	First, second, and third rows show images with different distortion types (distortion type is provided on the top of each image) taken from datasets with natural images, screen content images, and HDR-processed images, respectively. 
	 MOS and DMOS represent the subjective image quality scores;  
	  as DMOS (MOS) increases (decreases) the quality of the image degrades more.
	  Under each image we provide the predicted distortion type as well as  quality score computed via our proposed method.}
	\label{F1}
 \end{figure}

Objective quality metrics can be divided into full-reference (reference available or FR), reduced-reference (RR), and no-reference (reference not available or NR) methods based on the availability of a reference image \cite{wang2006modern}.
FR methods \cite{sheikh2005information,sheikh2006statistical,
kim2017deep,zhang2018unreasonable} usually provide the most precise evaluation results and perform well in predicting the quality scores of human subjects in comparison with RR and NR methods.
RR methods \cite{soundararajan2011rred,golestaneh2016reduced} provide a solution when the reference image is not fully accessible.
These methods generally operate by extracting a minimal set of parameters from the reference image; these parameters are later used with the distorted image to estimate quality.
However, in many practical applications, an IQA system does not have access to reference images. 
Without the reference image,   IQA task becomes very challenging.
The goal of the no-reference image quality assessment (NR-IQA) methods \cite{ghadiyaram2017perceptual,moorthy2011blind,
ma2017end,liang2016image,tang2011learning,tang2014blind,talebi2018nima, 
liu2017rankiqa,huang2018multitask,
mittal2012making,bosse2017deep,zhang2015som,ye2012unsupervised,
kang2014convolutional,ye2014beyond,saad2012blind,
wang2016distortion,xu2016multi,kang2015simultaneous,wu2017hierarchical,
kim2017fully,pan2018blind,lin2018hallucinated,mittal2012no,
zhang2018opinion,ma2017dipiq,
wu2019blind,fang2017no,gu2016learning,gu2017no,
yue2018analysis,min2017blind,kim2018deep,gu2014using,
wu2014no,li2019blind,zhang2015feature,
xue2013learning,
xue2014blind,he2012sparse,jiang2017blique,ying2019patches}
is to provide a solution when the reference image is not available.

NR-IQA methods are mainly divided into two groups, distortion-based and general-purpose methods. 
A distortion-based approach is to design NR algorithms for a specific type of distortion (e.g. blocking, blurring, or contrast distortions).
Distortion-based approaches have limited applications in more diverse scenarios.
A general-purpose approach is designed to evaluate image quality without being limited to distortion types.
General-Purpose methods usually make use of extracted features that are informative for various types of distortions.
Therefore, performance  highly depends on designing elaborate features. 
Existing general-purpose NR-IQA methods can mainly be classified into two categories depending on the types of features used.

The first category is based on well-chosen handcrafted features that are sensitive to the image quality (e.g. natural scene statistics, or image gradients).
Natural scene statistics (NSS) \cite{sheikh2005information,moorthy2011blind} is one of the most widely used features for IQA.
NSS has the assumption that natural images have statistical regularity that is altered when distortions are introduced.
Various types of NSS features have been defined in transformation domain  \cite{saad2012blind,saad2012blind,xue2014blind}, and spatial domain  \cite{mittal2012no,mittal2012making,zhang2015feature,li2019blind}.
The main constraint of NSS is its limitation in capturing and modeling the deviation among different similar distortions.
Moreover, the limitation of using handcrafted features is the lack of generalizability for modeling the multiple complex distortion types or contents.
As we will show in the results section, methods based on handcrafted features that are designed for natural images will not perform well for screen content images (SCIs) or high-dynamic-range- (HDR) processed images.

The second category is based on utilizing feature learning methods.
Inspired by the performance of  convolutional neural networks (CNNs) for different computer vision applications, different works utilize them for NR-IQA task.
The key factor behind achieving good performance via deep neural networks (DNNs) is having massive labeled datasets  \cite{deng2009imagenet} that can support the learning process \cite{krizhevsky2012imagenet}.
However, existing IQA datasets contain an extremely low number of labeled images.
Moreover, unlike generating datasets for image recognition task, generating large-scale reliable human subjective labels for quality assessment is very difficult.
Obtaining an IQA dataset requires a complex and time consuming psychometric experiment.
Furthermore, applying different data augmentation methods to increase the number of data can affect perceptual quality scores.
Nonetheless, different approaches such as transfer learning,    generative adversarial networks (GANs), and   proxy quality scores have been used to leverage the power of DNNs for NR-IQA.
Many researchers achieved state-of-the-art results by using DNNs for NR-IQA task.
With the exception of just a few number of algorithms  \cite{kang2015simultaneous,wang2016distortion,ma2017end,huang2018multitask,wu2014no,xu2016multi}, existing NR-IQA methods heavily rely only on the subjective quality scores to predict the quality.
Most of the learning-based methods ignore how utilizing  distortion during training can be beneficial to predict the perceptual quality,  similar to the way that HVS perceives the quality.

It is common to use multiple sources of information jointly in human learning.
Babies learn a new language by listening, speaking, and writing it. 
The problem of using a single network to solve multiple tasks has been repeatedly pursued in the context of deep
learning for computer vision.
Multi-task learning has achieved great performance for a variety of  vision tasks, such as  surface normal and depth prediction \cite{ren2018cross,eigen2015predicting}, object detection  \cite{peng2015learning}, and navigation \cite{zhu2017target}.

The perceptual process of the HVS includes multiple complex processes.
The visual sensitivity of the HVS varies according to different factors, such as distortion type, scene context, and spatial frequency of stimuli  \cite{daly1992visible,legge1980contrast,watson2005standard}. 
The HVS perceives image quality differently among various image contents  \cite{alam2014local,chou1995perceptually}.
Image quality assessment and distortion identification tasks are closely related.
During the feature learning process, identifying the distortion not only  helps the image quality assessment task, but also can open up the opportunity to enhance the quality of the distorted image based on the degradation type.
By leveraging the distortion type for the IQA task, our model  predicts the distortion type as well as the quality score of a distorted image during testing.
 Fig \ref{F1} shows example results of our proposed method.
 
From a modeling perspective, we are interested in answering:
 How does the additional distortion information influence NR-IQA and how much does it improve the overall performance? 
What  is the best architecture for taking advantage
of distortion information for NR-IQA?
We examine these questions empirically by evaluating
multiple DNN architectures that each take a different approach to combine information.
Our proposed method improves upon the following limitations of recent works on multi-task learning for NR-IQA  \cite{kang2015simultaneous,wang2016distortion,
ma2017end,huang2018multitask,wu2014no,xu2016multi}. 
\textbf{1)} 
The general trend among the recent multi-task NR-IQA methods 
\cite{kang2015simultaneous,wu2014no,wang2016distortion,ma2017end,huang2018multitask}  are to use overparameterized sub-networks or fully connected layers (FCL) for different tasks to achieve higher performance. 
In many real-world applications (e.g. self-driving cars and VR/AR)   IQA task needs to be carried out in a timely fashion on a computationally limited platform.
We propose a   pooling and fusion method along with $1\times1$ convolution layer which replace the overparameterized FCL for each task. 
\textbf{2)} Except for few methods,   multi-task NR-IQA methods
\cite{wu2014no,xu2016multi} use multi-stage training and optimize each task separately to achieve the best performance.
In contrast, our   method is simply trained end-to-end in one step without any multi-stage training.
\textbf{3)} 
Existing approaches without providing any analysis used the last layer of the network for all the tasks. 
They further use sub-networks or FCL for each task and rely on overparameterized learning layers to achieve good performance.
In this work, we empirically investigate the effect of feature fusion for NR-IQA task.
Using our design and feature fusion, we show that by using only  $1\times1$ convolution layers along with global average pooling (GAP) we can achieve a better performance.
From a computational perspective, by using a smaller backbone   in our experiments compared to existing models, our proposed model outperforms many of the existing state-of-the-art single-task and multi-task IQA algorithms.

We propose an end-to-end multi-task model, namely \textit{QualNet}, for NR-IQA.
During the training stage, our model makes use of distortion types as well as subjective human scores to predict the image quality.
We evaluate our approach against different NR-IQA methods and  achieve state-of-the-art results on several standard IQA datasets.
We  provide extensive experiments to demonstrate the effectiveness of our proposed method and architecture design.

In summary, our contributions  are summarized as follows:
\begin{itemize}

\item We propose an end-to-end  multi-task learning approach for blind quality assessment and distortion prediction.
We propose a feature fusion method to utilize distortion information for improving the quality score estimation task.
Specifically, given an input image, our proposed model is designed to regress the quality and predict the distortion type (or distortion types).
\item  We provide empirical experiments and evaluate different feature fusion choices to demonstrate the effectiveness of  our proposed model.

\item We  evaluate the performance of our proposed method on three well-known natural  image datasets (LIVE, CSIQ, TID2013). 
While using a smaller backbone for feature extraction, our model outperforms the existing algorithms on two datasets (CSIQ and TID2013).
We also test the performance of our method on the LIVE multi-distortion dataset and outperform the state-of-the-art NR-IQA methods.
In addition to natural image datasets, we further evaluate the performance of our algorithm on two other datasets with different scene domains (screen content images and HDR-processed images) and achieve state-of-the-art results.

\end{itemize}

\begin{figure*}[h]
\centering
  \includegraphics [scale=.25]{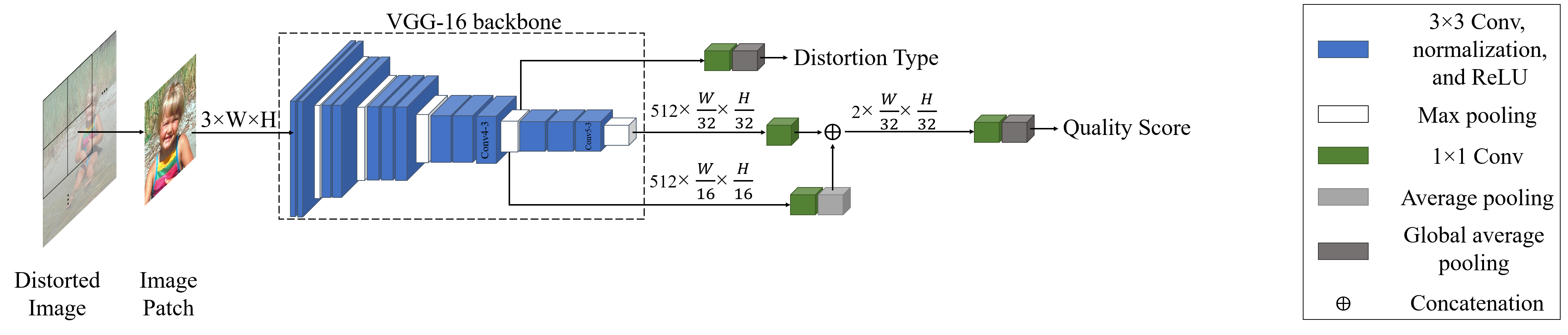}
	\caption{ Architecture of our proposed multi-task learning model. Conv indicates the convolutional layers, and normalization indicates instant normalization layers. }
	\label{F2}
 \end{figure*}

\section{Related Work}
We briefly review the literature related to our approach.

\subsection{Single-Task NR-IQAs}

Before the rise  of deep neural networks (DNNs), NR-IQA methods utilize different machine learning techniques (e.g.  dictionary learning) or image characteristics (e.g. NSS) to extract the features for predicting the quality score.
CORNIA \cite{ye2012unsupervised} used a dictionary learning method to encode the raw image patches to  features.
CORNIA features later were adopted in some NR-IQA models   \cite{ye2014beyond,zhang2015som,ma2017dipiq}.

With the progress of DNNs in different applications, more researchers have utilized them for NR-IQA.
In \cite{kang2014convolutional} the authors proposed a shallow CNN  for feature learning and quality regression.
 \cite{kim2018deep} developed a two-stage model that separated into an objective training stage followed by a subjective training stage.
In the first stage, they used PSNR to produce proxy scores.
Then, they generated the feature maps which were then regressed onto objective error maps.
The second stage aggregated the feature maps by weighted averaging and finally regressed these global features onto ground-truth subjective scores.
 \cite{pan2018blind} proposed a deep learning based model (BPSQM) which consists of a fully convolutional neural network and a deep pooling network.
Given a similarity index map, labels generated via a FR-IQA model, their model  produced a quality map that model the similarity index in pixel distortion level.
Hallucinated-IQA \cite{lin2018hallucinated} proposed a  NR-IQA method based on generative adversarial models.
They first generated a hallucinated reference image to compensate for the absence of the true reference.
Then, paired the information of hallucinated reference with the distorted image to estimate the quality score.
Although \cite{kim2018deep,pan2018blind,lin2018hallucinated} perform well for the NR-IQA task, they all used some sort of the reference image during their training  which  contradicts the NR-IQA purpose.

\subsection{Multi-Task NR-IQAs}
There are a few algorithms that attempt to do NR-IQA by leveraging the power of multi-task learning.
 \cite{kang2015simultaneous,wu2014no,wang2016distortion} designed a multi-task CNN to predict the type of distortions and image quality  from the last fully connected layer in the network.
 \cite{xu2016multi} developed a multi-task rank-learning-based IQA (MRLIQ) method.
They constructed multiple IQA models, each of which is responsible for one distortion type.
 \cite{ma2017end} proposed MEON, which is a multi-task network  where  two sub-networks train in two stages for distortion identification  and  quality prediction.
 \cite{huang2018multitask} proposed a model that used multi-task learning and dictionary learning for NR-IQA.

Among the aforementioned algorithms, 
 \cite{wu2014no,xu2016multi} do not train end-to-end and requires multi-stage training.
Although \cite{ma2017end} is an end-to-end method, the training process is performed in two
steps.
Existing methods mostly used just one fully connected layer for both tasks.
To the best of our knowledge, none of the existing works investigate the effect of feature fusion for NR-IQA task.
In this work, we investigate different feature fusion architectures to improve the performance of NR-IQA.
While taking advantage of multi-task learning, we use feature fusion from different blocks of the network to estimate the quality score more accurately.
Unlike the existing multi-task NR-IQA methods, we use global average pooling instead of fully connected layers which reduces learning parameters. 
 We observe that using fully connected layers can cause   overfitting and the network memorizes the training examples rather than generalizing from them. 

\section{Our Approach}
In this section, we introduce our proposed multi-task  model,   \textit{QualNet}, for NR-IQA.
\textit{QualNet} jointly learns distortion prediction as well as quality score prediction tasks.
Our proposed model is fully convolutional and is trained in an end-to-end manner.
An overview of the \textit{QualNet} architecture is given in Fig. \ref{F2}. 

\subsection{Problem Formulation}
Given a distorted image $I_d$, our goal is to estimate its distortion type (or distortion types) as well as quality score.
We partition the distorted image into overlapped patches $I_k$. 
Let $f_{\phi}$ represents proposed network with learnable parameters $\phi$.
Given an input image we have:
\begin{equation}
\boldsymbol{d},s=f_{\phi}({I}_{k})
\end{equation}
where ${I}_{k}$ is the input image to our network, and $s$ denotes the regressed quality score.
 $\boldsymbol{d}$ is a $m\times1$ vector, where $m$ indicates the total number of distortion types available in   dataset. 
The element of $\boldsymbol{d}$ with maximum value represents the index for the distortion type.
 
\subsection{Network architecture}
\textit{QualNet} is not limited by the choice of network architecture, any of state-of-the-art DNNs  \cite{simonyan2014very,he2016deep,szegedy2015going} can be used as a backbone for our proposed method.
However, in order to emphasize   the advantage of our method, we choose VGG16 which is a relatively smaller network compared to VGG19, Resnet34, or Resnet50, which are the ones that mostly used in the recent proposed NR-IQA methods  \cite{huang2018multitask,lin2018hallucinated,pan2018blind,wu2017hierarchical}.
Choosing a large network easily increases the number of parameters and makes the
network prone to overfitting instead of learning a better representation.

We show the architecture of our proposed model in Fig. \ref{F2}.
We modified VGG16 network to be used as the backbone.
We add instant normalization (IN) layer after each convolution layer.
However, in order to not increase the number of learnable parameters we set the trainable parameters for IN layers to be False. 
 Despite most of the existing deep learning based NR-IQA methods that use fully connected layers (FCL) for regressing the quality score, we did not use any fully connected layer in the head of our network, instead, we use global average pooling (GAP).
The use of GAP allows networks to function with less computational power and have better generalization performance.
In the head of our network, we use $1\times1$ convolution layer along with GAP layer  for each task.

The insight behind our network design comes from two observations.
\textbf{a)} Typically,   different FCL/sub-networks is used at the last layer of the backbone to separately compute each task.
We empirically observe that selecting features from the same layer of the network can reduce  the learning capability of that layer for both of the tasks.
That explains why previous models  needed additional overparameterized FCL/subnetwork in their architectures. 
 Therefore, here we choose features from different layers for each task.
\textbf{b)} Inspired by the work done in the field of understanding human visual perception and psychology, we know that human vision has different sensitivities to different levels and types of distortions  \cite{daly1992visible,legge1980contrast,watson2005standard}.
Thus, 
for   quality score prediction task it is important to take advantage of the features used to predict the type of distortion.
Fusion among features for tasks that are related can significantly capture the relative information and improve the performance by considering the relative information among all the different tasks.
We proposed to fuse the features used for distortion prediction task with the quality prediction features.
Using our proposed feature fusion, our method can perform efficiently and effectively well on different datasets without having overparameterized layers for each task.

\subsection{Distortion Prediction}
We use features of the max-pooling layer after $conv4{\_}3$ layer for the distortion prediction task. 
The output of max-pooling layer has $512\times\frac{W}{16}\times\frac{H}{16}$ dimension; $1\times1$ convolution along with GAP is used to compute the distortion type.
The output of $1\times1$ convolution has $m\times\frac{W}{16}\times\frac{H}{16}$ dimension, where $m$ is the total number of distortions available in the training data.
GAP layer is used to convert the $m\times\frac{W}{16}\times\frac{H}{16}$ feature map to $m\times1$ vector, which denotes by $\boldsymbol{d}$.
In the case of single distortion type, the element of $\boldsymbol{d}$ with maximum value represents the index for the distortion type.

\subsection{Quality Score Regression}
We use features of the max-pooling layer after $conv4{\_}3$ and $conv5{\_}3$ for the quality score regression task.
We first regress $conv4{\_}3$ and $conv5{\_}3$ features separately to obtain two coarse quality scores maps. 
The average pooling layer is used to make sure that the output of $1\times1$ convolutions after $conv4{\_}3$ has the same dimension (i.e. $1\times\frac{W}{32}\times\frac{H}{32}$) with the output of $1\times1$ convolutions after $conv5{\_}3$.
Finally, we combine the computed quality score maps by concatenating them and send it to a $1\times1$ convolution and GAP to achieve the final quality score. 

The HVS perceive image quality differently based on different distortion types; by concatenating the features from the layer that is used to predict the distortion type we observe that our model can utilize distortion type information for quality score prediction.
As shown in our ablation study, using our proposed feature fusion, we achieve the best performance.

\subsection{Training}
Here we describe the training details of the \textit{QualNet}.
Given $\textit{\textbf{d}}$ and $s$ as the output of our model, we use negative log-likelihood loss (which is simply a cross-entropy loss) and  $L2$ loss for optimizing distortion prediction and quality score regression tasks, respectively.
In other words, the total loss for our network is defined as:

\begin{subequations}
\begin{equation}
L_{total}=L_{d}+\lambda L_{s}
\end{equation}
\begin{equation}
L_{d}(\boldsymbol{d},c)=-\log(\frac{\exp(d_{c})}{\sum_{j=1}^{m}\exp(d_{j})})
\end{equation}
\begin{equation}
L_{s}=|s-g_{d}|^2
\end{equation}
\end{subequations}
where  $L_{d}$ and $L_{s}$ are the losses for distortion type prediction and quality score regression, respectively.
$\lambda$ is a regularization parameter that in our network is set to 1.
Eq. (2b) is the criterion that combines softmax and negative log-likelihood loss to train the distortion type classification problem with $m$ classes. 
In other words, $m$ is the number of distortion types.
$c$ denotes class index in the range $[1,m]$ as the target for the input.
$d$ is a $1 \times m$ vector which represents the output of the distortion type prediction task.
In Eq. (2c), $s$ and $g_d$ are regressed quality score and subjective human score for the image $I_d$, respectively.

In \textit{QualNet} framework, the sizes of input images must be fixed to train the model on a GPU. 
Therefore, each input image should be divided into multiple patches of the same size. 
In our experiment, we choose  patch size of $128\times128$, 
we set the step of the sliding window to $64$, i.e. the neighboring patches are overlapped by
$64$ pixels.
We consider the patch size large enough to
reflect the overall image quality, we set the quality score of
each patch to its distorted images subjective ground-truth score. 
The effect of different patch sizes is provided in our ablation study.
To expand the training
data set, the horizontal flip is performed in our training process for data augmentation.

For both tasks, our network is optimized end-to-end simultaneously.
The proposed network is trained iteratively via backpropagation over a number of 50 epochs.
We set batch size to 1.
For optimization, we use the adaptive moment
estimation optimizer (ADAM)  \cite{kingma2014adam} with
$\beta_1$ = 0.9, $\beta_2$ = 0.999, we set the initially learning rate to $2\times10^{-4}$.
We set the learning rate decay to $0.98$, and it applied after every 3 epochs.
We fine-tune our model end-to-end while using pretrained Imagenet weights to initialize the weights of the VGG16 network, the rest of the weights in the network are randomly initialized.
Similar to the existing NR-IQA models for all of our evaluations, to train and test
the \textit{\textit{QualNet}}, we randomly divide the reference images into
two subsets, $80\%$ for training and $20\%$ for testing. Then,
the corresponding distorted images are divided into training and testing sets so that there are no overlaps between
the two. 
All the experiments are under
ten times random train-test splitting operation, and the median SROCC and LCC values are reported as final statistics.

\section{Results}

In this section, the performance of our proposed model is analyzed in terms of its ability to predict subjective ratings of image quality as well as distortion type. 
We evaluate the performance of our proposed model extensively.
We use seven standard image quality datasets for our performance evaluation.
For a distorted image $I_d$, the final predicted quality score is simply defined by averaging the predicted quality scores over all the patches   from $I_d$.
Also, the final image distortion
is decided by a majority voting of the patches belong to $I_d$, i.e. the most frequently occurring distortion on patches determines the distortion of the image.

First, we study the effectiveness of our proposed model in regards to its ability to predict the image quality in a manner that agrees with subjective perception.
For performance evaluation, we employ two commonly used performance metrics. 
We measure the prediction monotonicity of \textit{QualNet} via the Spearman rank-order correlation coefficient (SROCC).
This metric operates only on the rank of the data points and ignores the relative distance between data points.
We also apply  regression analysis to provide a nonlinear mapping between the objective scores and either the subjective mean opinion scores (MOS) or difference of mean opinion scores (DMOS).
We measure the Pearson linear correlation coefficient (LCC) between MOS (DMOS) and the objective scores after nonlinear regression.

We further provide accuracy of \textit{QualNet} for the distortion type prediction task.
Finally, we provide ablation studies to evaluate the performance of \textit{QualNet} for 
different choices of architecture, fusion, patch sizes, and optimization strategy.
%
\subsection{Datasets}
The detailed information for   datasets that we use for our evaluation is summarized in Table \ref{TB1}.
Specifically, we perform experiments on seven widely used
benchmark datasets.
For natural images  use LIVE  \cite{sheikh2006statistical}, CSIQ  \cite{larson2010most}, TID2008  \cite{ponomarenko2009tid2008},
TID2013  \cite{ponomarenko2013color}, LIVE-MD  \cite{jayaraman2012objective}.
For SCIs we use SIQAD \cite{yang2015perceptual}, and for images with different tone-mapping, multi-exposure fusion, and post-processing we use ESPL-LIVE HDR  \cite{kundu2017large}.

\begin{table}[h]
\centering
\caption{Summary of the datasets evaluated in our experiments.}
\resizebox{3.2 in}{!} {
\begin{tabular}{ccccc}
\hline 
\multirow{2}{*}{Databases} & \# of Dist. & \# of Dist.  & Multiple Distortions  & \multirow{2}{*}{Score Type}\tabularnewline
 & Images & Types & per images? & \tabularnewline
\hline 
LIVE & 799 & 5 & NO & DMOS\tabularnewline
CSIQ & 866 & 6 & NO & DMOS\tabularnewline
TID2008 & 1700 & 17 & NO & MOS\tabularnewline
TID2013 & 3000 & 24 & NO & MOS\tabularnewline
LIVE-MD1 & 255 & 2 & YES & DMOS\tabularnewline
LIVE-MD2 & 255 & 2 & YES & DMOS\tabularnewline
SIQAD & 980 & 7 & NO & DMOS\tabularnewline
ESPL-LIVE HDR & 1811 & 11 & NO & MOS\tabularnewline
\hline 
\end{tabular} }
\label{TB1}
\end{table}

\subsection{Natural Images}

\begin{table*}[h]
\centering
\caption{Comparison of \textit{QualNet} vs. various NR-IQA algorithms on different datasets. Bold entries are the best and second-best performers.}
\resizebox{6.7 in}{!} {
\begin{tabular}{c|cc|cc|cc|cc|cc|cc}
\multicolumn{1}{c}{} &  & \multicolumn{1}{c}{} &  & \multicolumn{1}{c}{} &  & \multicolumn{1}{c}{} &  & \multicolumn{1}{c}{} &  & \multicolumn{1}{c}{} &  & \tabularnewline
\hline 
\hline 
\multirow{2}{*}{Method} & \multicolumn{2}{c|}{LIVE} & \multicolumn{2}{c|}{CSIQ} & \multicolumn{2}{c|}{TID2013} & \multicolumn{2}{c|}{LIVE-MD1} & \multicolumn{2}{c|}{LIVE-MD2} & \multicolumn{2}{c}{Average}\tabularnewline
\cline{2-13} \cline{3-13} \cline{4-13} \cline{5-13} \cline{6-13} \cline{7-13} \cline{8-13} \cline{9-13} \cline{10-13} \cline{11-13} \cline{12-13} \cline{13-13} 
 & SROCC & LCC & SROCC & LCC & SROCC & LCC & SROCC & LCC & SROCC & LCC & SROCC & LCC\tabularnewline
\hline 
DIIVINE\cite{saad2012blind} & 0.892 & 0.908 & 0.804 & 0.776 & 0.643 & 0.567 & 0.909 & 0.931 & 0.831 & 0.874 & 0.815 & 0.811\tabularnewline
BRISQUE\cite{mittal2012no} & 0.929 & 0.944 & 0.812 & 0.748 & 0.626 & 0.571 & 0.904 & 0.936 & 0.861 & 0.888 & 0.826 & 0.817\tabularnewline
NIQE\cite{mittal2012making} & 0.908 & 0.948 & 0.812 & 0.629 & 0.421 & 0.330 & 0.861 & 0.911 & 0.782 & 0.844 & 0.765 & 0.732\tabularnewline
IL-NIQE\cite{zhang2015feature} & 0.902 & 0.906 & 0.822 & 0.865 & 0.521 & 0.648 & 0.881 & 0.857 & 0.871 & 0.869 & 0.799 & 0.829\tabularnewline
BIECON\cite{kim2017fully} & 0.958 & 0.961 & 0.815 & 0.823 & 0.717 & 0.762 & - & - & - & - & 0.830 & 0.848\tabularnewline
IQA-CNN++ \cite{kang2015simultaneous} & 0.965 & 0.966 & \textbf{0.892} & 0.905 & 0.872 & 0.878 & \textbf{0.953} & 0.942 & 0.943 & 0.905 & \textbf{0.925} & 0.919\tabularnewline
CNN-SWA\cite{wang2016distortion} & \textbf{0.982} & 0.974 & 0.887 & 0.891 & \textbf{0.880} & 0.851 & 0.934 & 0.921 & 0.931 & 0.921 & 0.922 & 0.891\tabularnewline
MEON\cite{ma2017end} & 0.951 & 0.955 & 0.852 & 0.864 & 0.808 & 0.824 & 0.915 & 0.934 & \textbf{0.953} & \textbf{0.949} & 0.895 & 0.905\tabularnewline
HFD-BIQA\cite{wu2017hierarchical} & 0.951 & 0.972 & 0.842 & 0.890 & 0.764 & 0.681 & - & - & - & - & 0.852 & 0.847\tabularnewline
WaDIQaM-NR\cite{bosse2017deep} & 0.960 & 0.955 & 0.852 & 0.844 & 0.835 & 0.855 & - & - & - & - & 0.882 & 0.884\tabularnewline
BPSQM\cite{pan2018blind} & 0.973 & 0.963 & 0.874 & 0.915 & 0.862 & \textbf{0.885} & 0.867 & 0.898 & 0.891 & 0.912 & 0.893 & 0.914\tabularnewline
Hallucinated-IQA\cite{lin2018hallucinated} & \textbf{0.982} & \textbf{0.982} & 0.885 & 0.910 & 0.879 & 0.880 & - & - & - & - & 0.915 & 0.924\tabularnewline
DIQA \cite{kim2018deep} & 0.975 & 0.977 & 0.884 & 0.915 & 0.825 & 0.850 & 0.945 & \textbf{0.951} & 0.932 & 0.944 & 0.912 & \textbf{0.927}\tabularnewline
Ref. \cite{huang2018multitask} & 0.970 & 0.971 & 0.889 & 0.894 & 0.862 & 0.884 & 0.927 & 0.926 & 0.932 & 0.939 & 0.916 & 0.922\tabularnewline
NRVPD\cite{li2019blind} & 0.956 & 0.960 & 0.886 & \textbf{0.918} & 0.749 & 0.808 & 0.937 & 0.942 & 0.924 & 0.941 & 0.890 & 0.913\tabularnewline
QualNet (proposed) & \textbf{0.980} & \textbf{0.984} & \textbf{0.907} & \textbf{0.921} & \textbf{0.890} & \textbf{0.901} & \textbf{0.961} & \textbf{0.965} & \textbf{0.960} & \textbf{0.952} & \textbf{0.938} & \textbf{0.943}\tabularnewline
\hline 
\end{tabular}
}
\label{TB2}
\end{table*}

Most of the existing NR-IQA designed to predict the quality of natural images.
Table \ref{TB2}
shows the obtained performance evaluation results of our proposed algorithm on the LIVE, CSIQ, TID2013, LIVE-MD1, and LIVE-MD2 datasets in comparison with state-of-the-art general-purpose NR-IQA algorithms.
As shown in Table \ref{TB2}, our proposed method outperforms state-of-the-art algorithms on several datasets while having a smaller backbone.
We believe that this improvement is because of our feature fusing and taking advantage of multi-task learning.
Although  \cite{lin2018hallucinated} achieved the best performance for SROCC on LIVE dataset, it has  bigger backbone compared to us (VGG19 vs VGG16).
Moreover, as shown in Table \ref{TB2}, our proposed model achieves the highest performance when we average the performances among all the datasets.

%
\textbf{ Cross-dataset evaluations.} To evaluate the generalizability of the \textit{QualNet}, we conduct
cross dataset test.
Training is performed on
LIVE, and then the obtained model is tested on TID2008 (for comparability)
 without parameter adaptation.
 Both  quality score regression and  distortion type prediction tasks are tested.
  We follow the common experiment setting to test the results on the subsets of
TID2008, where four distortion types (i.e., JPEG, JPEG2K,
WN, and Blur) are included, and   logistic regression is
applied to match the predicted DMOS to MOS value  \cite{rohaly2000video,sheikh2006statistical}.

\begin{table}[h]
\centering
\caption{SROCC and LCC comparison of various NR-IQA models trained using LIVE dataset and tested on the TID2008 dataset. Bold entries are the best and second-best performers.}
\resizebox{2.0 in}{!} {
\begin{tabular}{l|c|c}
\multicolumn{1}{l}{} & \multicolumn{1}{c}{} & \tabularnewline
\hline 
\hline 
Methods & SROCC & LCC\tabularnewline
\hline 
CORNIA \cite{ye2012unsupervised} & 0.880 & 0.890\tabularnewline
CNN \cite{kang2014convolutional} & 0.920 & 0.903\tabularnewline
SOM \cite{zhang2015som} & 0.923 & 0.899\tabularnewline
IQA-CNN++ \cite{kang2015simultaneous} & 0.917 & 0.921\tabularnewline
CNN-SWA \cite{wang2016distortion} & 0.915 & 0.922\tabularnewline
dipIQ \cite{ma2017dipiq} & 0.916 & 0.918\tabularnewline
MEON \cite{ma2017end} & 0.921 & 0.918\tabularnewline
WaDIQaM-NR \cite{bosse2017deep} & 0.919 & 0.916\tabularnewline
DIQA \cite{kim2018deep} & 0.922 & -\tabularnewline
BPSQM \cite{pan2018blind} & 0.910 & -\tabularnewline
HIQA \cite{lin2018hallucinated} & \textbf{0.934} & 0.917\tabularnewline
Ref. \cite{huang2018multitask} & \textbf{0.935} & \textbf{0.936}\tabularnewline
NRVPD \cite{li2019blind} & 0.904 & 0.908\tabularnewline
QualNet (Proposed) & 0.925 & \textbf{0.940}\tabularnewline
\hline 
\multicolumn{1}{l}{} & \multicolumn{1}{c}{} & \tabularnewline
\end{tabular}

}
\label{TB3}
\end{table}

The results provided in Table \ref{TB3} demonstrate the generalization ability of
our approach.
As shown in Table \ref{TB3}, \textit{QualNet} outperforms all  existing algorithms in terms of LCC.
It also achieves comparable results in terms of SROCC.
Although  \cite{huang2018multitask} and  \cite{lin2018hallucinated} achieved higher results compared to our method, it worth mentioning that they both use more learning parameters in their models.
  \cite{huang2018multitask} and  \cite{lin2018hallucinated} used Resnet50 and VGG19 as their backbones, respectively.
For the distortion type prediction task, \textit{QualNet} predicted the distortion types of TID2008 images with $92\%$ accuracy  while trained on LIVE images.

\begin{figure}[h]
\centering
  \includegraphics [scale=.095]{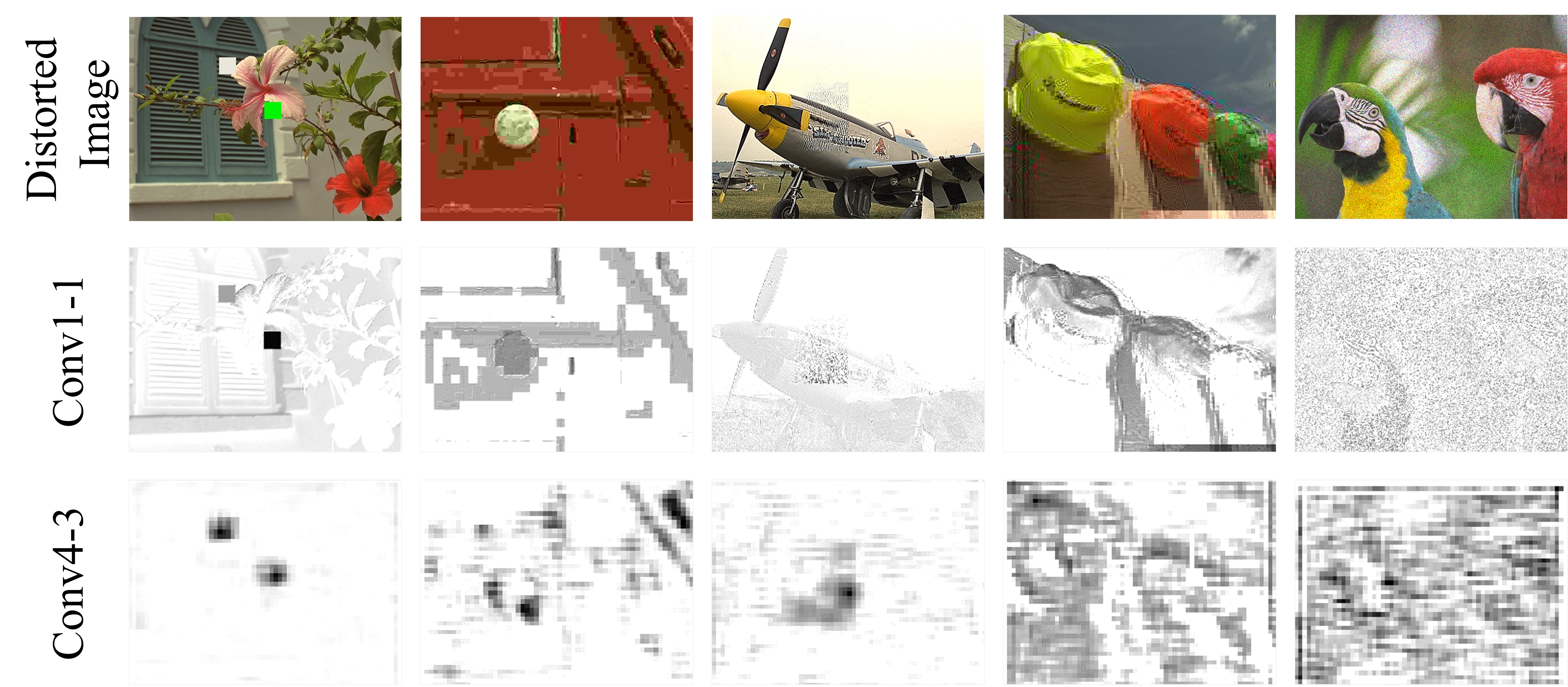}
	\caption{ Illustration of the feature maps of  the $conv1{\_}1$ block (2nd row) and $conv4{\_}3$ block (3rd row) for images with different distortion types.
	Distorted images (1st row)  are taken from TID2013 dataset.
From first to fifth column the images have JPEG, JPEG2000, JPEG transmission errors, and Gaussian noise distortions, respectively.
The images in the third row are resized for illustration purposes. }
	\label{F3}
 \end{figure}

To validate if our proposed method is consistent with human visual perception, we visualize the feature maps of $conv1{\_}1$ and $conv4{\_}3$ blocks in Fig. \ref{F3}.
The first row in Fig. \ref{F3}
 shows five different distortion types from TID2013, including Local block-wise distortions of different intensity (first column),  JPEG (second column), JPEG2000 (third column), JPEG transmission errors
(fourth column), and Gaussian noise (fifth column).
The second and third rows correspond to the feature maps of $conv1{\_}1$ and $conv4{\_}3$ blocks, respectively.
In contrast to recent methods that used the reference images to teach their networks to focus on distorted areas and generate a quality map,  Fig. \ref{F3} shows that our method automatically learns the distortions and highlight them. 
 The
dark areas in images in the second and third rows in Fig. \ref{F3} indicate distorted regions.
We  observe that our proposed model learns to focus on the distortions instead of the content of   images.
For instance, the last row of Fig. \ref{F3} clearly shows that for a noisy image our method captures the noise artifacts instead of the image texture/content.

\subsection{Screen Content Images}
Most of the NR-IQA algorithms were developed for natural images and they do not typically perform well on SCIs.
Here we show that our proposed method not only performs well on natural images, but it also works well for images with other contents.

Unlike natural images, SCIs include diverse forms of visual
 content, such as pictorial and textual regions. 
Therefore, the characteristics of SCIs and those of
 natural images are greatly different.
 Recently, there have been some metrics that designed specifically for the visual quality prediction of SCIs  \cite{gu2016learning,gu2017no,fang2017no,wu2019blind}. 
 Here we use SIQAD dataset to evaluate the performance of \textit{QualNet} on SCIs.
 The SIQAD dataset includes 980 screen content images corrupted by conventional  distortion types (e.g. JPEG, blur, noise, etc).
Table \ref{T4} provides a comparison between our results and various modern NR-IQA algorithms designed either for natural images  \cite{mittal2012making,zhang2015feature,kang2015simultaneous,bosse2017deep,huang2018multitask} or specifically for SCIs  \cite{gu2016learning,gu2017no,fang2017no,wu2019blind}. 
The results show that our algorithm yields a high correlation with the subjective quality ratings and yields the best results in terms of both LCC and SROCC.

\begin{table}[t]
\centering
\caption{Comparison of \textit{QualNet} vs. various NR-IQA algorithms on SIQAD
 dataset. Bold entries are the best and second-best performers.}
\resizebox{2.2 in}{!} {
\begin{tabular}{lcc}
 & & \tabularnewline
\hline 
\hline 
Database & \multicolumn{2}{c}{SIQAD}\tabularnewline
\hline 
Methods & SROCC & LCC\tabularnewline
\hline 
\hline 
NIQE  \cite{mittal2012making} & 0.482 & 0.500\tabularnewline
IL-NIQE  \cite{zhang2015feature} & 0.517 & 0.540\tabularnewline
IQA-CNN++  \cite{kang2015simultaneous} & 0.702 & 0.721\tabularnewline
CNN-SWA  \cite{wang2016distortion} & 0.725 & 0.735\tabularnewline
BMS  \cite{gu2016learning} & 0.725 & 0.756\tabularnewline
ASIQE  \cite{gu2017no} & 0.757 & 0.788\tabularnewline
NRLT  \cite{fang2017no} & 0.820 & 0.844\tabularnewline
WaDIQaM-NR  \cite{bosse2017deep} & \textbf{0.852} & \textbf{0.859}\tabularnewline
Ref.  \cite{wu2019blind} & 0.811 & 0.833\tabularnewline
Ref.  \cite{huang2018multitask} & 0.844 & 0.856\tabularnewline
QualNet (Proposed) & \textbf{0.853} & \textbf{0.862}\tabularnewline
\hline 
\end{tabular}
}
\label{T4}
\end{table}
\subsection{HDR-processed images}

There is a growing practice of acquiring/creating and displaying high dynamic range (HDR) images and other types of pictures created by multiple exposure fusion. 
These kinds of images allow for more pleasing representation and better use of the available luminance and color ranges in real scenes, which can range from direct sunlight to faint starlight  \cite{reinhard2010high}.
HDR images typically are obtained by blending a stack of Standard Dynamic Range (SDR) images at varying exposure levels, HDR images need to be tone-mapped to SDR for display on standard monitors.
 Multi Exposure Fusion (MEF) techniques are also used to bypass HDR creation by fusing an exposure stack directly to SDR images to achieve aesthetically pleasing luminance and color distributions.
 HDR images may also be post-processed (color saturation, color temperature, detail enhancement, etc.) for aesthetic purposes.
Therefore, due to different types of tone mapping, multi-exposure fusion, and post-processing techniques, HDR images can go under different types of distortions that are different from conventional    distortions.

To demonstrate the effectiveness of \textit{QualNet} and its application for NR-IQA in different domains we conduct  an experiment where we evaluate  the performance of several state-of-the-art NR-IQA algorithms on the recently developed ESPL-LIVE HDR dataset.
 The images in the ESPL-LIVE HDR dataset were  obtained  using  11  HDR  processing  algorithms involving both tone-mapping and MEF. 
ESPL-LIVE HDR dataset  also considered post-processing artifacts of HDR image creation, which typically occur in commercial HDR systems.

\begin{table}[t]
\centering
\caption{Comparison of \textit{QualNet} vs. various NR-IQA algorithms on ESPL-LIVE HDR
 dataset. Bold entries are the best and second-best performers.}
\resizebox{2.25 in}{!} {
\begin{tabular}{lcc}
 & & \tabularnewline
\hline 
\hline 
Database & \multicolumn{2}{c}{ESPL-LIVE HDR}\tabularnewline
\hline 
Methods & SROCC & LCC\tabularnewline
\hline 
\hline 
DIIVINE  \cite{saad2012blind} & 0.523 & 0.530\tabularnewline
GM-LOG  \cite{xue2014blind} & 0.549 & 0.562\tabularnewline
IQA-CNN++  \cite{kang2015simultaneous} & 0.673 & 0.685\tabularnewline
CNN-SWA  \cite{wang2016distortion} & 0.66 & 0.672\tabularnewline
Ref.  \cite{huang2018multitask} & 0.701 & 0.695\tabularnewline
BTMQI  \cite{gu2016blind} & 0.668 & 0.673\tabularnewline
WaDIQaM-NR  \cite{bosse2017deep} & 0.752 & 0.762\tabularnewline
BLIQUE-TMI  \cite{jiang2017blique} & 0.704 & 0.712\tabularnewline
HIGRADE  \cite{kundu2017no} & 0.695 & 0.696\tabularnewline
Ref.  \cite{chen2019blind} & \textbf{0.763} & \textbf{0.768}\tabularnewline
QualNet (Proposed) & \textbf{0.796} & \textbf{0.786}\tabularnewline
\hline 
\end{tabular}
}
\label{T5}
\end{table}
Table \ref{T5} provides a comparison between our results and various modern NR-IQA algorithms designed either for natural images  \cite{saad2012blind,kang2015simultaneous,xue2014blind,wang2016distortion,huang2018multitask,bosse2017deep} or specifically for HDR-processed images \cite{chen2019blind,kundu2017no,jiang2017blique,gu2016blind}.
 The results show that our algorithm outperforms existing algorithms in terms of both SROCC and LCC and yields a high correlation with the subjective quality ratings.

\subsection{Distortion Prediction}
Although the main focus of this paper is on NR-IQA, in Table \ref{T6} we provide the results of the distortion type prediction task for our proposed model.
As shown in Table \ref{T6}, \textit{QualNet}  achieves good prediction accuracy over different datasets.

\begin{table}[h]
\centering
\caption{Performance of our proposed model for distortion type prediction task on different datasets.}
\resizebox{3.3 in}{!} {
\begin{tabular}{c|cccc}
\multicolumn{1}{c}{} & & & & \tabularnewline
\hline 
\hline 
datasets & LIVE & CSIQ & TID2008 & TID2013\tabularnewline
\hline 
Distortion Prediction (\%) & 83 & 93 & 89 & 91\tabularnewline
\hline 
datasets & LIVE-MD1 & LIVE-MD2 & SIQAD & ESPL-LIVE HDR\tabularnewline
\hline 
Distortion Prediction (\%) & 97 & 96 & 98 & 68\tabularnewline
\hline 
\hline 
\multicolumn{1}{c}{} & & & & \tabularnewline
\end{tabular}

}
\label{T6}
\end{table}
\subsection{Ablation Studies}
To investigate the effectiveness of our module and training
scheme, we provide a comprehensive ablation study in this section.
For each ablated model we train it on  the subsets of the LIVE dataset and test it on the subsets of
TID2013, where four distortion types (i.e., JPEG, JPEG2K,
WN, and Blur) are included.

\begin{figure}[t]
\centering
  \includegraphics [scale=.15]{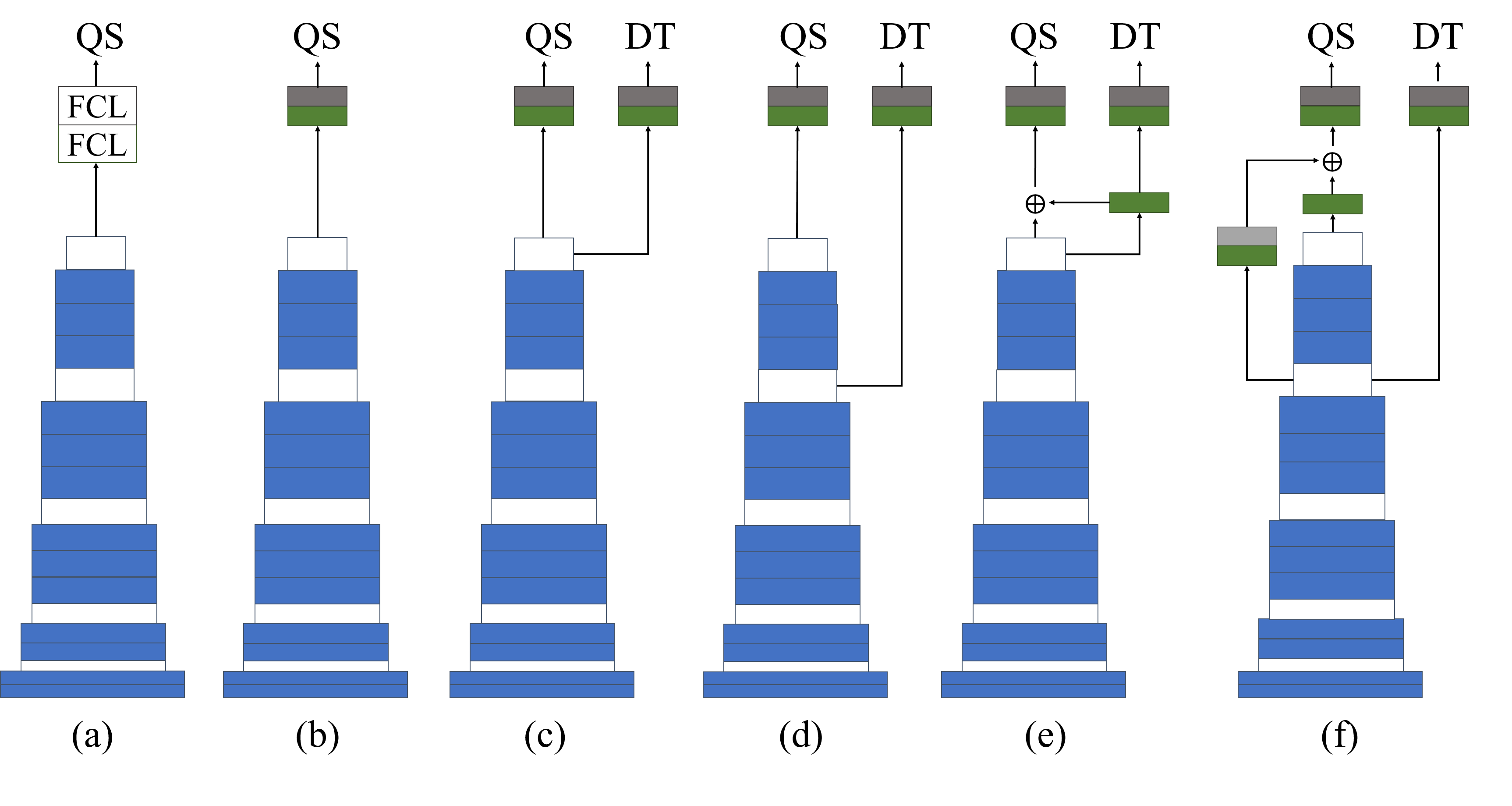}
	\caption{ Exploring different network architectures for   NR-IQA task.
	QS and DT stand for quality score and distortion type, respectively.
	(a) and (b)  show   single-task models, where they are using FCL  and $1\times1$ convolution along with GAP,   respectively, in the last layer for quality score estimation.
	(c)-(f) show a multi-task model while having different architectures.}
	\label{F4}
 \end{figure}

Fig. \ref{F4} demonstrates different network architectures for our ablation study.
Fig. \ref{F4} (a) shows a model with just quality score regression task (single-task) while using fully connected layers for the regression task; 
Fig. \ref{F4} (b) is the same as Fig. \ref{F4} (a) while replacing the fully connected layers with $1\times1$ convolution along with GAP.
Fig. \ref{F4} (c) shows the model in a multi-task manner, but both quality score and distortion type prediction are done by using the features from the last layer without any feature fusion.
Fig. \ref{F4} (d) shows the model in a multi-task manner, while the quality score and distortion prediction are done by using the features from different layers in the network without any feature fusion.
Fig. \ref{F4} (e) shows the model in a multi-task manner, while feature fusion is performed from the last layer for quality estimation.
Fig. \ref{F4} (f) is our proposed method.

Table \ref{T7} shows the results of different ablated models from Fig. \ref{F4}.
As we can see model-a and model-b achieve the lowest performance comparing to the multi-task models. 
This proves our hypothesis that using distortion type knowledge along with quality score can improve the results.
Moreover, we observe that model-b outperforms model-a; 
in our experiments, we observe that using fully connected layers while having limited training data will cause the model to overfit to the training data quickly which causes a lack of generality in the test phase.
Furthermore, as shown in Table \ref{T7}, both model-c and model-d achieve very similar results, but worse than model-f. 
This demonstrates the effectiveness of our feature fusion method. 
Finally, we observe that using the features from different layers for each task leads to   better performance. 
In Table \ref{T7} we also provide the performance evaluation of our model via SGD for optimization,
We observe that using SGD can cause the network to converge slower and lead to   slightly worse results.
Finally we provide results of our model while using different patch sizes as an input.
We can see that while patch size of 128 leads to the best results as we move to 64 and 32 the performance degrades more. 
The main reason for dropping the performance while using smaller patch sizes is that the small patches do not have enough content information to represent the ground truth subjective score of the distorted image.
In this paper, we select VGG16 as our backbone to show the effectiveness of our model   to other methods that chose deeper backbones.
However, as shown in Table \ref{T7} (Model-f-VGG-19) using a deeper network (e.g. VGG19) can improve our results even more.

%
\section{Conclusion}
In this paper, we proposed a simple yet effective multi-task model, \textit{QualNet},  for general-purpose no-reference image quality assessment (NR-IQA). 
Our model exploits distortion type as well as subjective human scores.  
We demonstrate that by employing multi-task learning
as well as our proposed feature fusion method, our model achieves better performance across different datasets.
Our experimental results show that the proposed model achieves high accuracy while maintaining consistency with human perceptual quality assessments.

\begin{table}[t]
\centering
\caption{Ablation study results}
\resizebox{2.85 in}{!} {
\begin{tabular}{c|c|c|c|c}
\multicolumn{1}{c}{} & \multicolumn{1}{c}{} & \multicolumn{1}{c}{} & \multicolumn{1}{c}{} & \tabularnewline
\hline 
\hline 
Methods & Optmization & Patch Size & SROCC & LCC\tabularnewline
\hline 
\hline 
Model-a & ADAM & 128$\times$128 & 0.862 & 0.882\tabularnewline
Model-b & ADAM & 128$\times$128 & 0.870 & 0.902\tabularnewline
Model-c & ADAM & 128$\times$128 & 0.905 & 0.918\tabularnewline
Model-d & ADAM & 128$\times$128 & 0.902 & 0.922\tabularnewline
Model-e & ADAM & 128$\times$128 & 0.906 & 0.920\tabularnewline
Model-f (Proposed) & ADAM & 128$\times$128 & 0.916 & 0.936\tabularnewline
Model-f-V2 & SGD & 128$\times$128 & 0.909 & 0.929\tabularnewline
Model-f-V3 & ADAM & 64$\times$64 & 0.891 & 0.911\tabularnewline
Model-f-V4 & ADAM & 32$\times$32 & 0.878 & 0.889\tabularnewline
Model-f-VGG-19 & ADAM & 128$\times$128 & 0.925 & 0.945\tabularnewline
\hline 
\multicolumn{1}{c}{} & \multicolumn{1}{c}{} & \multicolumn{1}{c}{} & \multicolumn{1}{c}{} & \tabularnewline
\end{tabular} 
}
\label{T7}
\end{table}

\bibliographystyle{ACM-Reference-Format}
\bibliography{samplebase}

\end{document}